\documentclass[11pt,a4paper]{article}

\usepackage{amsmath}            
\usepackage{booktabs}
\usepackage[font={small,it},justification=centering]{caption}
\usepackage{coling2016}
\usepackage{graphicx}           
\usepackage{latexsym}
\usepackage{multicol}           
\usepackage{multirow}
\usepackage{paralist}
\usepackage{tabularx}
\usepackage{times}
\usepackage{url}

\newcommand{\F}[0]{$F_1$}

\newcommand{\ttranslate}[2]{#1

\noindent\emph{#2}}

\title{Generating Sentiment Lexicons for German Twitter}

\author{Uladzimir Sidarenka \and Manfred Stede\\
  Applied Computational Linguistics\\
  UFS Cognitive Science\\
  University of Potsdam / Germany\\
  {\tt \{sidarenk|stede\}@uni-potsdam.de}}

\date{}

\begin{document}

\maketitle
\begin{abstract}
  Despite substantial progress made in developing new sentiment
  lexicon generation (SLG) methods for English, the task of
  transferring these approaches to other languages and domains in a
  sound way still remains open.  In this paper, we contribute to the
  solution of this problem by systematically comparing semi-automatic
  translations of common English polarity lists with the results of
  the original automatic SLG algorithms, which were applied directly
  to German data.  We evaluate these lexicons on a corpus of 7,992
  manually annotated tweets.  In addition to that, we also collate the
  results of dictionary- and corpus-based SLG methods in order to find
  out which of these paradigms is better suited for the inherently
  noisy domain of social media.  Our experiments show that
  semi-automatic translations notably outperform automatic systems
  (reaching a macro-averaged \F{}-score of 0.589), and that
  dictionary-based techniques produce much better polarity lists as
  compared to corpus-based approaches (whose best \F{}-scores run up
  to 0.479 and 0.419 respectively) even for the non-standard Twitter
  genre.  All reimplementations of the compared systems and the
  resulting lexicons of these methods are available online at
  {\small\url{https://github.com/WladimirSidorenko/SentiLex}}.
\end{abstract}

%
%
\blfootnote{






    \hspace{-0.65cm}  
    This work is licenced under a Creative Commons
    Attribution 4.0 International License.
    License details:
    \url{http://creativecommons.org/licenses/by/4.0/}
}

\section{Introduction}\label{sec:intro}

Sentiment lexicons play a crucial role in many existing and emerging
opinion mining applications.  Not only do they serve as a valuable
source of features for supervised classifiers
\cite{Mohammad:13,Zhu:14} but they also achieve competitive results
when used as the main component of a sentiment analysis system
\cite{Taboada:11}.  Due to this high impact and tremendous costs of
building such lexicons manually, devising new algorithms for an
automatic generation of polarity lists has always been an area of
active research in the sentiment analysis literature
\cite[pp.~79-91]{Liu:12}.  Nevertheless, despite some obvious progress
in this field \cite{Cambria:16b}, the applicability of these
approaches to other languages and text genres still raises questions:
It is, for instance, unclear whether simply translating the existing
English sentiment resources would produce better results than applying
the methods that were initially proposed for their creation directly
to the target language.  Furthermore, for automatic systems which draw
their knowledge from lexical taxonomies, such as \textsc{WordNet}
\cite{Miller:95}, it remains unanswered whether these approaches would
also work for languages in which such resources are much smaller in
size, and, even if they would, whether the resulting lexicons would
then be general enough to carry over to more colloquial texts.
Finally, for methods which derive their polarity lists from text
corpora, it is not clear whether these approaches would still yield an
acceptable quality when operating on inherently noisy input data.

In this paper, we try to analyze these and other problems in detail,
using the example of German Twitter.  More precisely, given a
collection of German microblogs with manually labeled polar terms and
prior polarities of these expressions, we want to find an SLG method
that can best predict these terms and their semantic orientation.  For
this purpose, we compare the existing German sentiment lexicons (most
of which were semi-automatically translated from popular English
resources) with the results of common automatic dictionary- and
corpus-based SLG approaches.


We begin our study by describing the data set which will be used in
our evaluation. Afterwards, in Section~\ref{sec:eval-metrics}, we
introduce the metrics with which we will assess the quality of various
polarity lists.  Then, in Section~\ref{sec:semi-automatic}, we
evaluate three most popular existing German sentiment lexicons---the
German Polarity Clues \cite{Waltinger:10}, SentiWS \cite{Remus:10},
and Zurich Polarity List of \newcite{Clematide:10}, subsequently
comparing them with popular automatic SLG approaches in
Section~\ref{sec:automatic}.  Finally, after estimating the impact of
different seed sets on the automatic methods and performing a
qualitative analysis of their entries, we draw our conclusions and
outline directions for future research in the final part of this
paper.

To avoid unnecessary repetitions, we deliberately omit a summary of
related work, since most of the popular SLG algorithms will be
referenced in the respective evaluation sections anyway.  We should,
however, note that, apart from the research on the automatic lexicon
generation, our study is also closely related to the experiments of
\newcite{Andreevskaia:08} and the ``Sentiment Analysis in Twitter''
track of the SemEval competition
\cite{Nakov:13,Rosenthal:14,Rosenthal:15}.  In contrast to the former
work, however, where the authors trained a supervised classifier on
one domain and applied it to another in order to determine the
polarities of the sentences, we \emph{explicitly model a situation
  where no annotated training data are available}, thus looking for
the most general unsupervised SLG strategy which performs best
regardless of the target domain, and we also \emph{evaluate these
  strategies on the level of lexical phrases only}.  Furthermore,
unlike in the SemEval track, where the organizers also provided
participants with sufficient labeled in-domain training sets and then
asked them to predict the contextual polarity of pre-annotated polar
expressions in the test data, we \emph{simultaneously try to predict
  polar terms and their prior polarities, learning both of them
  without supervision}.

\section{Data}\label{sec:data}

We perform our evaluation on the publicly available Potsdam Twitter
Sentiment corpus (PotTS; Sidarenka, 2016).\footnote{We use version
  0.1.0 of this corpus.}  This collection comprises 7,992 microblogs
pertaining to the German federal elections, general political life,
papal conclave~2013, as well as casual everyday conversations.  Two
human experts annotated these posts with polar terms and their prior
polarities,\footnote{The annotators had been asked to judge the
  semantic orientation of a term irrespective of its possible
  negations.  They could, however, consider the context for
  determining whether a particular reading of a polysemous word in the
  text was subjective or not.} reaching a substantial agreement of
0.75 binary $\kappa$ \cite{Cohen:60}.\footnote{A detailed
  inter-annotator agreement study of this corpus is provided in
  \cite{Sidarenka:16}.} We used the complete data set labeled by one
of the annotators as our test corpus, getting a total of 6,040
positive and 3,055 negative terms including multi-word expressions.
However, since many of these expressions were emoticons, which, on the
one hand, were a priori absent in common lexical taxonomies due to
their colloquial nature and therefore not amenable to dictionary-based
SLG systems but, on the other hand, could be easily captured by
regular expressions, we decided to exclude non-alphabetic smileys
altogether from our study.  This left us with a set of 3,459 positive
and 2,755 negative labeled terms (1,738 and 1,943 unique expressions
respectively), whose $\kappa$-agreement run up to 0.59.  Besides the
test set, we selected a small subset of 400 tweets from the other
annotator and used it as development data for tuning the
hyper-parameters of the tested approaches.\footnote{That way, we only
  used the labeled corpus for evaluation or parameter optimization,
  other resources---\textsc{GermaNet} \cite{Hamp:97} and the German
  Twitter Snapshot \cite{Scheffler:14}---were used for training the
  methods.}

\section{Evaluation Metrics}\label{sec:eval-metrics}

A central question to our experiments are the evaluation metrics that
we should use for measuring lexi\-con quality.  Usually, this quality is
estimated either \textit{intrinsically} (i.e., taking a lexicon in
isolation and immediately assessing its accuracy) or
\textit{extrinsically} (i.e., considering the lexicon within the scope
of a bigger application such as a supervised classifier which utilizes
lexicon's entries as features).

Traditionally, intrinsic evaluation of English sentiment lexicons
amounts to comparing these polarity lists with the General Inquirer
(GI; Stone, 1966)---a manually compiled set of 11,895 words annotated
with their semantic categories---by taking the intersection of the two
resources and estimating the percentage of matches in which
automatically induced polar terms have the same polarity as the GI
entries.  This evaluation method, however, is somewhat problematic:
First of all, it is not easily transferable to other languages, since
even a manual translation of the GI lexicon is not guaranteed to cover
all language- and domain-specific polar expressions.  Secondly, due to
the intersection, this method does not penalize for a low recall so
that a lexicon consisting of just two terms \textit{good}$^+$ and
\textit{bad}$^-$ will have the highest possible score, often
surpassing polarity lists with a greater number of entries.  Finally,
this comparison does not account for polysemy.  As a result, an
ambiguous word only one of whose (possibly rare) senses is subjective
will always be ranked the same as a purely polar term.

Unfortunately, an extrinsic evaluation does not always provide a
solution in this case, since, depending on the type of the extrinsic
system (e.g., a document classifier), it might still presuppose a
large data set for training other system components and, furthermore,
might yield overly high scores, which, however, are mainly due to
these extrinsic modules rather than the quality of the lexicons
themselves.

Instead of using these approaches, we opt for a direct comparison of
the induced polarity lists with an existing annotated corpus, since
this type of evaluation allows us to solve at least three of the
previously mentioned issues: It does account for the recall, it does
accommodate polysemous words,\footnote{Recall that the annotators of
  the PotTS data set were asked to annotate a polar expression iff its
  actual sense in the respective context was polar.} and it does
preclude intermediate components which might artificially boost the
results.  In particular, in order to check a lexicon against the PotTS
data set, we construct a case-insensitive trie
\cite[pp. 492--512]{Knuth:98} from the lexicon entries and match this
trie against the contiguously running corpus text,\footnote{In other
  words, we successively compare lexicon entries with the occurrences
  of corpus tokens in the same linear order as these occurrences
  appear in the text.}  simultaneously comparing it with the actual
word forms and lemmas of corpus tokens.\footnote{We use the
  \textsc{TreeTagger} of \newcite{Schmid:95} for lemmatization.}  A
match is considered correct iff the matched entry absolutely
corresponds to the (possibly lemmatized) expert's annotation and has
the same polarity as the one specified by the human coder.  That way,
we estimate the precision, recall, and \F{}-score for each particular
polarity class (positive, negative, and neutral), considering all
words absent in the lexicons (not annotated in the corpus) as neutral.

\section{Semi-Automatic Lexicons}\label{sec:semi-automatic}

We first apply the above metric to estimate the quality of the
existing German resources: the German Polarity Clues (GPC; Waltinger,
2010), SentiWS (SWS; Remus, 2010), and the Zurich Polarity List (ZPL)
of \newcite{Clematide:10}.

The GPC set comprises 10,141 subjective entries automatically
translated from the English sentiment lexicons Subjectivity Clues
\cite{Wilson:05} and SentiSpin \cite{Takamura:05}, with a subsequent
manual correction of these translations, and several synonyms and
negated terms added by the authors.  The SWS lexicon includes 1,818
positively and 1,650 negatively connoted terms, also providing their
part-of-speech tags and inflections (resulting in a total of 32,734
word forms).  Similarly to the GPC, the authors used an English
sentiment resource---the GI lexicon of \newcite{Stone:66}---to
bootstrap their polarity list, manually revising these automatic
translations afterwards.  In addition to that, \newcite{Remus:10} also
expanded their set with words and phrases frequently co-occurring with
positive and negative seed lexemes using collocation information
obtained from a corpus of 10,200 customer reviews and the German
Collocation Dictionary \cite{Quasthoff:10}.  Finally, the Zurich
Polarity List features 8,000 subjective entries taken from
\textsc{GermaNet} synsets \cite{Hamp:97}.  These synsets were manually
annotated with their prior polarities by human experts.  Since the
authors, however, found the number of polar adjectives obtained that
way insufficient for running further classification experiments, they
automatically enriched this lexicon with more attributive terms by
analyzing conjoined corpus collocations using the method of
\newcite{Hatzivassi:97}.

\begin{table}[h]
  \begin{center}
    \bgroup \setlength\tabcolsep{0.1\tabcolsep}\scriptsize
    \begin{tabular}{p{0.162\columnwidth} 
        *{9}{>{\centering\arraybackslash}p{0.074\columnwidth}} 
        *{2}{>{\centering\arraybackslash}p{0.068\columnwidth}}} 
      \toprule
      \multirow{2}*{\bfseries Lexicon} & %
      \multicolumn{3}{c}{\bfseries Positive Expressions} & %
      \multicolumn{3}{c}{\bfseries Negative Expressions} & %
      \multicolumn{3}{c}{\bfseries Neutral Terms} & %
      \multirow{2}{0.068\columnwidth}{\bfseries\centering Macro\newline \F{}} & %
      \multirow{2}{0.068\columnwidth}{\bfseries\centering Micro\newline \F{}}\\
      \cmidrule(lr){2-4}\cmidrule(lr){5-7}\cmidrule(lr){8-10}

      & Precision & Recall & \F{} & %
      Precision & Recall & \F{} & %
      Precision & Recall & \F{} & & \\\midrule


      GPC & 0.209 & 0.535 & 0.301 & %
      0.195 & 0.466 & 0.275 & %
      0.983 & 0.923 & 0.952 & %
      0.509 & 0.906 \\


      SWS & 0.335 & 0.435 & 0.379 & %
      0.484 & 0.344 & \textbf{0.402} & %
      0.977 & 0.975 & 0.976 & %
      0.586 & 0.952\\


      ZPL & 0.411 & 0.424 & 0.417 & %
      0.38 & 0.352 & 0.366 & %
      0.977 & 0.979 & 0.978 & %
      0.587 & 0.955 \\


      GPC $\cap$ SWS $\cap$ ZPL & \textbf{0.527} & 0.372 & \textbf{0.436} & %
      \textbf{0.618} & 0.244 & 0.35 & %
      0.973 & \textbf{0.99} & \textbf{0.982} & %
      \textbf{0.589} & \textbf{0.964} \\


      GPC $\cup$ SWS $\cup$ ZPL & 0.202 & \textbf{0.562} & 0.297 & %
      0.195 & \textbf{0.532} & 0.286 & %
      \textbf{0.985} & 0.917 & 0.95 & %
      0.51 & 0.901 \\\bottomrule
    \end{tabular}
    \egroup
    \caption{Evaluation of semi-automatic German sentiment lexicons.\\
      {\small GPC -- German Polarity Clues \cite{Waltinger:10}, SWS --
        SentiWS \cite{Remus:10}, ZPL -- Zurich Polarity Lexicon
        \cite{Clematide:10}}}
    \label{snt-lex:tbl:gsl-res}
  \end{center}
\end{table}

For our evaluation, we tested the three lexicons in isolation and also
built their union and intersection in order to check for ``synergy''
effects.  The results are shown in Table~\ref{snt-lex:tbl:gsl-res}.
As can be seen from the statistics, with a few exceptions, the highest
scores for all classes as well as the best macro- and micro-averaged
\F{}-measures are achieved by the intersection of all three lexicons.
On the other hand, as expected, the highest recall of polar
expressions (and consequently the best precision at recognizing
neutral terms) is attained by the union of these resources.  The only
case where individual lexicons are able to outperform these
combinations is observed for the \F{}-score of the negative class,
where both SentiWS and ZPL show better results than their
intersection, which is mainly due to the higher recall of these two
polarity lists.

\section{Automatic Methods}\label{sec:automatic}

A natural question which arises upon the evaluation of the existing
semi-automatic resources is how well fully automatic methods can
perform in comparison with these lexicons.  Traditionally, automatic
SLG algorithms have been grouped into dictionary- and corpus-based
ones, with their own complementary strengths and weaknesses.
Dictionary-based approaches, for instance, incorporate distilled
linguistic knowledge from a typically manually labeled lexical
database, but lack any domain specificity.  Corpus-based methods, on
the other hand, can operate directly on unannotated in-domain data,
but often have to deal with an extreme noisiness of their input.
Since it was unclear which of these properties would have a stronger
impact on the net results, we decided to reimplement the most commonly
used algorithms from both of these paradigms and evaluate them on the
PotTS corpus.

\subsection{Dictionary-Based Approaches}\label{ssec:dba}

For dictionary-based methods, we adopted the systems proposed by
\newcite{Hu:04}, \newcite{Blair-Goldensohn:08}, \newcite{Kim:04},
\newcite{Esuli:06c}, as well as the min-cut and label-propagation
approaches of \newcite{Rao:09}, and the random-walk algorithm
described by \newcite{Awadallah:10}.

The first of these works \cite{Hu:04} expanded a given set of seed
terms with known semantic orientations by propagating polarity values
of these terms to their \textsc{WordNet} synonyms and passing reversed
polarity scores to the antonyms of these words.  Later on, this idea
was further refined by \newcite{Blair-Goldensohn:08}, who obtained
polarity labels for new terms by multiplying a score vector $\vec{v}$
containing the orientation scores of the known seed words (-1 for
negative expressions and 1 for positive ones) with an adjacency matrix
$A$ constructed for the \textsc{WordNet} graph.  
With various modifications, the core idea of passing the polarity
values through a lexical graph was adopted in almost all of the
following dictionary-based works: \newcite{Kim:04}, for instance,
computed the polarity class for a new word $w$ by multiplying the
prior probability of this class with the likelihood of the word $w$
occurring among the synonyms of the seed terms with the given semantic
orientation, choosing at the end the polarity which maximized this
equation.  Other ways of bootstrapping polarity lists were proposed by
\newcite{Esuli:06c}, who created their \textsc{SentiWordNet} resource
using a committee of Rocchio and SVM classifiers trained on
successively expanded sets of polar terms; \newcite{Rao:09}, who
adopted the min-cut approach of \newcite{Blum:04}, also comparing it
with the label-propagation algorithm of \newcite{Zhu:02}; and,
finally, \newcite{Awadallah:10}, who used a random walk method by
estimating the polarity of an unknown word as the difference between
an average number of steps a random walker had to make in order to
reach a term from the positive or negative set.

Since some of these approaches relied on different seed sets or
pursued different objectives (two- versus three-way classification),
we decided to unify their settings and interfaces for the sake of our
experiments.  In particular, we were using the same translated seed
list of \newcite{Turney:03} for all methods, expanding this set by 10
neutral terms (``neutral'' \emph{neutral}, ``sachlich''
\emph{objective}, ``technisch'' \emph{technical}, ``finanziell''
\emph{financial} etc.).\footnote{All translated seed sets are provided
  along with the source code for this paper.}  Additionally, we
enhanced all binary systems to ternary classifiers, so that each
tested method could differentiate between positive, negative, and
neutral terms.  In the final step, we applied these methods to
\textsc{GermaNet} \cite{Hamp:97}---a German equivalent of the English
\textsc{WordNet} \cite{Miller:95}, which, however, is much smaller in
size, having 20,792 less synsets for the three common parts of speech
(nouns, adjectives, and verbs) than the Princeton resource.

\begin{table}[h]
  \begin{center}
    \bgroup \setlength\tabcolsep{0.1\tabcolsep}\scriptsize
    \begin{tabular}{p{0.142\columnwidth} 
        >{\centering\arraybackslash}p{0.06\columnwidth} 
        *{9}{>{\centering\arraybackslash}p{0.072\columnwidth}} 
        *{2}{>{\centering\arraybackslash}p{0.058\columnwidth}}} 
      \toprule
      \multirow{2}*{\bfseries Lexicon} & %
      \multirow{2}*{\bfseries \# of Terms} & %
      \multicolumn{3}{c}{\bfseries Positive Expressions} & %
      \multicolumn{3}{c}{\bfseries Negative Expressions} & %
      \multicolumn{3}{c}{\bfseries Neutral Terms} & %
      \multirow{2}{0.068\columnwidth}{\bfseries\centering Macro\newline \F{}} & %
      \multirow{2}{0.068\columnwidth}{\bfseries\centering Micro\newline \F{}}\\
      \cmidrule(lr){3-5}\cmidrule(lr){6-8}\cmidrule(lr){9-11}

      & & Precision & Recall & \F{} & %
      Precision & Recall & \F{} & %
      Precision & Recall & \F{} & & \\\midrule


      \textsc{Seed Set} & 20 & \textbf{0.771} & 0.102 & 0.18 & %
      0.568 & 0.017 & 0.033 & %
      0.963 & \textbf{0.999} & \textbf{0.981} & %
      0.398 & \textbf{0.962}\\


      HL & 5,745 & 0.161 & 0.266 & 0.2 & %
      0.2 & 0.133 & 0.16 & %
      0.969 & 0.96 & 0.965 & %
      0.442 & 0.93\\


      BG & 1,895 & 0.503 & 0.232 & \textbf{0.318} & %
      0.285 & 0.093 & 0.14 & %
      0.968 & 0.991 & 0.979 & %
      \textbf{0.479} & 0.959\\


      KH & 356 & 0.716 & 0.159 & 0.261 & %
      0.269 & 0.044 & 0.076 & %
      0.965 & 0.997 & \textbf{0.981} & %
      0.439 & \textbf{0.962}\\


      ES & 39,181 & 0.042 & \textbf{0.564} & 0.078 & %
      0.033 & \textbf{0.255} & 0.059 & %
      \textbf{0.981} & 0.689 & 0.81 & %
      0.315 & 0.644\\


      RR$_{\textrm{mincut}}$ & 8,060 & 0.07 & 0.422 & 0.12 & %
      0.216 & 0.073 & 0.109 & %
      0.972 & 0.873 & 0.92 & %
      0.383 & 0.849\\


      RR$_{\textrm{lbl-prop}}$ & 1,105 & 0.567 & 0.176 & 0.269 & %
      \textbf{0.571} & 0.046 & 0.085 & %
      0.965 & 0.997 & \textbf{0.981} & %
      0.445 & \textbf{0.962}\\


      AR & 23 & 0.768 & 0.1 & 0.176 & %
      0.568 & 0.017 & 0.033 & %
      0.963 & \textbf{0.999} & \textbf{0.981} & %
      0.397 & \textbf{0.962}\\


      HL $\cap$ BG $\cap$ RR$_{\textrm{lbl-prop}}$ & 752 & 0.601 & 0.165 & 0.259 & %
      0.567 & 0.045 & 0.084 & %
      0.965 & 0.997 & \textbf{0.981} & %
      0.441 & \textbf{0.962}\\


      HL $\cup$ BG $\cup$ RR$_{\textrm{lbl-prop}}$ & 6,258 & 0.166 & 0.288 & 0.21 & %
      0.191 & 0.146 & \textbf{0.165} & %
      0.97 & 0.958 & 0.964 & %
      0.446 & 0.929\\\bottomrule
    \end{tabular}
    \egroup
    \caption{Evaluation of dictionary-based approaches.\\ {\small HL
        -- \newcite{Hu:04}, BG -- \newcite{Blair-Goldensohn:08}, KH --
        \newcite{Kim:04}, ES -- \newcite{Esuli:06c}, RR --
        \newcite{Rao:09}, AR -- \newcite{Awadallah:10}}}
    \label{snt-lex:tbl:lex-res}
  \end{center}
\end{table}

The results of this evaluation are shown in
Table~\ref{snt-lex:tbl:lex-res}.  This time, the situation is much
more varied, as different systems can achieve best results on just
some aspects of certain classes but can hardly attain best overall
scores in all categories.  This is, for instance, the case for the
positive and negative polarities, where the best precision scores are
reached by the seed set in the first case and the label propagation
algorithm of \newcite{Rao:09} in the second case.  However, with
respect to the recall, both of these polarity lists perform notably
worse than the approach of \newcite{Esuli:06c}.  Yet other
systems---the matrix-vector method of \newcite{Blair-Goldensohn:08}
and the union of the three overall top-scoring systems
respectively---reach the highest \F{}-scores for these two classes.
Nevertheless, we can still notice three main tendencies in this
evaluation:
\begin{inparaenum}[\itshape i\upshape)]
\item the method of \newcite{Esuli:06c} generally gets the highest
  recall of polar terms and, consequently, achieves the best precision
  in recognizing neutral words, but suffers from a low precision for
  the positive and negative polarities;
\item simultaneously five systems attain the same best \F{}-scores on
  recognizing neutral terms, which, in turn, leads to the best
  micro-averaged \F{}-results for all polarity classes; and, finally,
\item the system of \newcite{Blair-Goldensohn:08} shows the best
  macro-averaged performance.  This approach, however, is extremely
  susceptible to its hyper-parameter settings (in particular, we
  considered the maximum number of times the initial vector $\vec{v}$
  was multiplied with the adjacency matrix $A$ as such a parameter and
  noticed a dramatic decrease of method's scores after the fifth
  iteration).
\end{inparaenum}

\subsection{Corpus-Based Approaches}\label{ssec:cba}

An alternative way to generate polarity lists is to use corpus-based
approaches.  In contrast to dictionary-based methods, these systems
typically operate immediately on raw texts and are, therefore,
virtually independent of any manually annotated linguistic resources.
This flexibility, however, might come at the cost of a reduced
accuracy due to an inherent noisiness of the unlabeled data.  The most
prominent representatives of this class of algorithms are the
approaches proposed by \newcite{Takamura:05}, \newcite{Velikovich:10},
\newcite{Kiritchenko:14}, and \newcite{Severyn:15}, which we briefly
describe in this section.

Drawing on the pioneering work of \newcite{Hatzivassi:97}, in which
the authors expanded an initial list of polar adjectives by analyzing
coordinately conjoined terms from a text corpus, \newcite{Takamura:05}
enhanced this algorithm, extending it to other parts of speech and
also incorporating semantic links from \textsc{WordNet} in addition to
the co-occurrence statistics extracted from the corpus.  After
representing the final set of terms as an electron lattice, whose edge
weights corresponded to the contextual and semantic links between
words, the authors computed the most probable polarity distribution
for this lattice by adopting the Ising spin model from statistical
mechanics.

The approach of \newcite{Velikovich:10} was mainly inspired by the
label-propagation algorithm of \newcite{Rao:09}, with the crucial
difference that, instead of taking an averaged sum of the adjacent
neighbor values when propagating the label scores through the graph,
the authors took the maximum of these scores in order to prune
unreliable, noisy corpus links.  Similarly, \newcite{Kiritchenko:14}
built on the method of \newcite{Turney:03} and computed polarity
scores for new words by taking the difference of their PMI
associations with noisy labeled positive and negative classes.
Finally, \newcite{Severyn:15} trained a supervised SVM classifier on a
distantly labeled data set and included the top-ranked unigram and
bigram features in their final lexicon.

For our evaluation, we applied these methods to the German Twitter
Snapshot \cite{Scheffler:14}---a collection of 24~M microblogs
gathered in April, 2013, constructing the collocation graph from the
lemmatized word forms of this corpus and only considering words which
appeared at least four times in the analyzed data.  We again were
using the \textsc{TreeTagger} of \newcite{Schmid:95} for lemmatization
and \textsc{GermaNet} \cite{Hamp:97} for deriving semantic links
between word vertices for the method of \newcite{Takamura:05}.

\begin{table}[h]
  \begin{center}
    \bgroup \setlength\tabcolsep{0.1\tabcolsep}\scriptsize
    \begin{tabular}{p{0.142\columnwidth} 
        >{\centering\arraybackslash}p{0.06\columnwidth} 
        *{9}{>{\centering\arraybackslash}p{0.072\columnwidth}} 
        *{2}{>{\centering\arraybackslash}p{0.058\columnwidth}}} 
      \toprule
      \multirow{2}*{\bfseries Lexicon} & %
      \multirow{2}*{\bfseries \# of Terms} & %
      \multicolumn{3}{c}{\bfseries Positive Expressions} & %
      \multicolumn{3}{c}{\bfseries Negative Expressions} & %
      \multicolumn{3}{c}{\bfseries Neutral Terms} & %
      \multirow{2}{0.068\columnwidth}{\bfseries\centering Macro\newline \F{}} & %
      \multirow{2}{0.068\columnwidth}{\bfseries\centering Micro\newline \F{}}\\
      \cmidrule(lr){3-5}\cmidrule(lr){6-8}\cmidrule(lr){9-11}

      & & Precision & Recall & \F{} & %
      Precision & Recall & \F{} & %
      Precision & Recall & \F{} & & \\\midrule


      \textsc{Seed Set} & 20 & \textbf{0.771} & 0.102 & 0.18 & %
      \textbf{0.568} & 0.017 & 0.033 & %
      0.963 & \textbf{0.999} & \textbf{0.981} & %
      0.398 & \textbf{0.962}\\


      TKM & 920 & 0.646 & \textbf{0.134} & \textbf{0.221} & %
      0.565 & \textbf{0.029} & \textbf{0.055} & %
      \textbf{0.964} & 0.998 & \textbf{0.981} & %
      \textbf{0.419} & \textbf{0.962}\\


      VEL & 60 & 0.764 & 0.102 & 0.18 & %
      \textbf{0.568} & 0.017 & 0.033 & %
      0.963 & 0.999 & 0.98 & %
      0.398 & \textbf{0.962}\\


      KIR & 320 & 0.386 & 0.106 & 0.166 & %
      \textbf{0.568} & 0.017 & 0.033 & %
      0.963 & 0.996 & 0.979 & %
      0.393 & 0.959\\


      SEV & 60 & 0.68 & 0.102 & 0.177 & %
      \textbf{0.568} & 0.017 & 0.033 & %
      0.963 & \textbf{0.999} & \textbf{0.981} & %
      0.397 & \textbf{0.962}\\

      TKM $\cap$ VEL $\cap$ SEV & 20 & \textbf{0.771} & 0.102 & 0.18 & %
      \textbf{0.568} & 0.017 & 0.033 & %
      0.963 & \textbf{0.999} & \textbf{0.981} & %
      0.398 & \textbf{0.962}\\


      TKM $\cup$ VEL $\cup$ SEV & 1,020 & 0.593 & \textbf{0.134} & 0.218  & %
      0.565 & \textbf{0.029} & \textbf{0.055} & %
      \textbf{0.964} & 0.998 & 0.98 & %
      0.418 & \textbf{0.962}\\\bottomrule
    \end{tabular}
    \egroup
    \caption{Evaluation of corpus-based approaches.\\ {\small TKM --
        \newcite{Takamura:05}, VEL -- \newcite{Velikovich:10}, KIR --
        \newcite{Kiritchenko:14}, SEV -- \newcite{Severyn:15}}}
    \label{snt-lex:tbl:corp-meth}
  \end{center}
\end{table}

The results of these experiments are shown in
Table~\ref{snt-lex:tbl:corp-meth}.  This time, we can observe a clear
superiority of Takamura et al.'s method, which not only achieves the
best recall and \F{} in recognizing positive and negative items but
also attains the highest micro- and macro-averaged results for all
three polarity classes.
The cardinality of the other induced lexicons, however, is much
smaller than the size of Takamura et al.'s polarity list.  Moreover,
these lexicons also show absolutely identical scores for the negative
expressions as the original seed set.  Since these results were
somewhat unexpected, we decided to investigate the reasons for
possible problems.  As it turned out, the macro-averaged \F{}-values
of these methods were rapidly going down on the held-out development
set as the number of their induced polar terms increased.  Since we
considered the lexicon size as one of the hyper-parameters of the
tested approaches, we immediately stopped populating these lexicons
when we noticed a decrease in their results.  As a consequence, only
the highest-ranked terms (all of which had the positive polarity) were
included in the final lists.

One of the reasons for such rapid quality decrease was the
surprisingly high positive bias of the initial seed set: While
converting the original seed list of \newcite{Turney:03} to German, we
translated the English word ``correct'' as ``richtig''.  This German
word, however, also has another reading which means \emph{real} (as in
\emph{a real fact} or \emph{a real sports car}) and which was much
more frequent in the analyzed snapshot, often appearing in an
unequivocally negative context, e.g., ``ein richtiger Bombenanschlag''
(\emph{a real bomb attack}) or ``ein richtiger Terrorist'' (\emph{a
  real terrorist}).  As a consequence of this, methods relying on
distant supervision had to deal with an extremely unbalanced training
set (the automatically labeled corpus that we distantly obtained for
the approach of \newcite{Kiritchenko:14} using these seeds, for
instance, had 716,210 positive versus 92,592 negative training
instances).

\section{Effect of Seed Sets}\label{sec:seedsets}

Since the set of the initial seed terms appeared to play an important
role for at least three of the tested methods, we decided to analyze
the impact of this factor in more detail by repeating our experiments
with the seed lists proposed by \newcite{Hu:04}, \newcite{Kim:04},
\newcite{Esuli:06c}, and \newcite{Remus:10}.  For this purpose, we
manually translated the seed sets of \newcite{Hu:04} and
\newcite{Kim:04} into German.  Since the authors, however, only
provided some examples of their seeds without specifying the full
lists, we filled up our translations with additional polar terms to
match the original cardinalities.  A different procedure was applied
to obtain the seed set of \newcite{Esuli:06c}---since this resource
comprised a vast number of neutral terms (the authors considered as
neutral all words from the General Inquirer lexicon which were not
marked there as either positive or negative), we automatically
translated the neutral subset of these seeds with the help of a
publicly available translation site
({\small\url{http://www.dict.cc}}), using the first suggestion
returned by this service for each original English term.

\begin{figure}[hbtp!]
  \centering
  \includegraphics[height=12em,width=0.8\linewidth]{%
    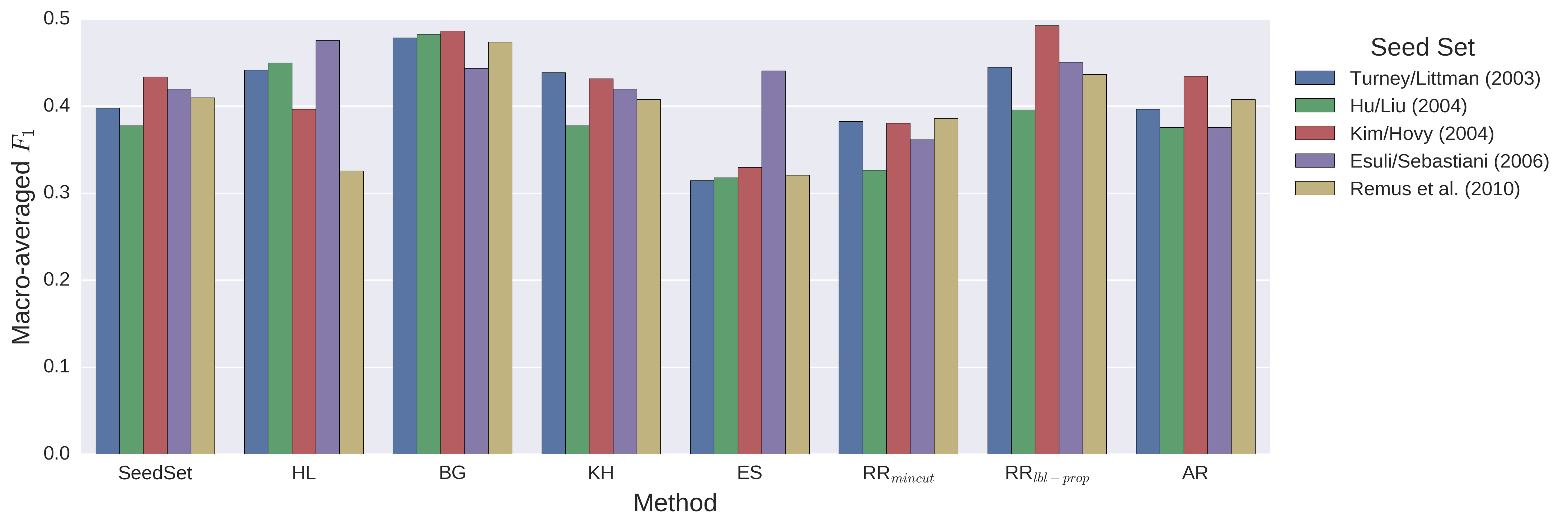}
  \caption{Macro-averaged \F{}-scores of the dictionary-based approaches
    with different seed sets.}\label{snt:fig:sent-dict-lex-alt-seeds}
\end{figure}

The updated results for the dictionary-based approaches with the
alternative seed sets are shown in
Figure~\ref{snt:fig:sent-dict-lex-alt-seeds}.  This time, we again can
notice superior scores achieved by the method of
\newcite{Blair-Goldensohn:08}, which not only performs better than the
other systems on average but also seems to be less susceptible to the
varying quality and size of the different seed lists.  The remaining
methods typically achieve their best macro-averaged results with
either of the two top-scoring polarity sets---the seed list of
\newcite{Kim:04} or the seed set of \newcite{Esuli:06c}.  This is, for
instance, the case for the method of \newcite{Kim:04} and the min-cut
approach of \newcite{Rao:09}, whose performance with the native
Kim-Hovy seed set is on par with their results achieved using the
Turney-Littman seeds.  The label-propagation and random walk
algorithms can even strongly benefit from the seeds provided by
\newcite{Kim:04}.  The remaining two methods---\newcite{Hu:04} and
\newcite{Esuli:06c}---work best in combination with the initial
polarity set proposed by \newcite{Esuli:06c}.

\begin{figure}[hbtp!]
  \centering
  \includegraphics[height=12em,width=0.8\linewidth]{%
    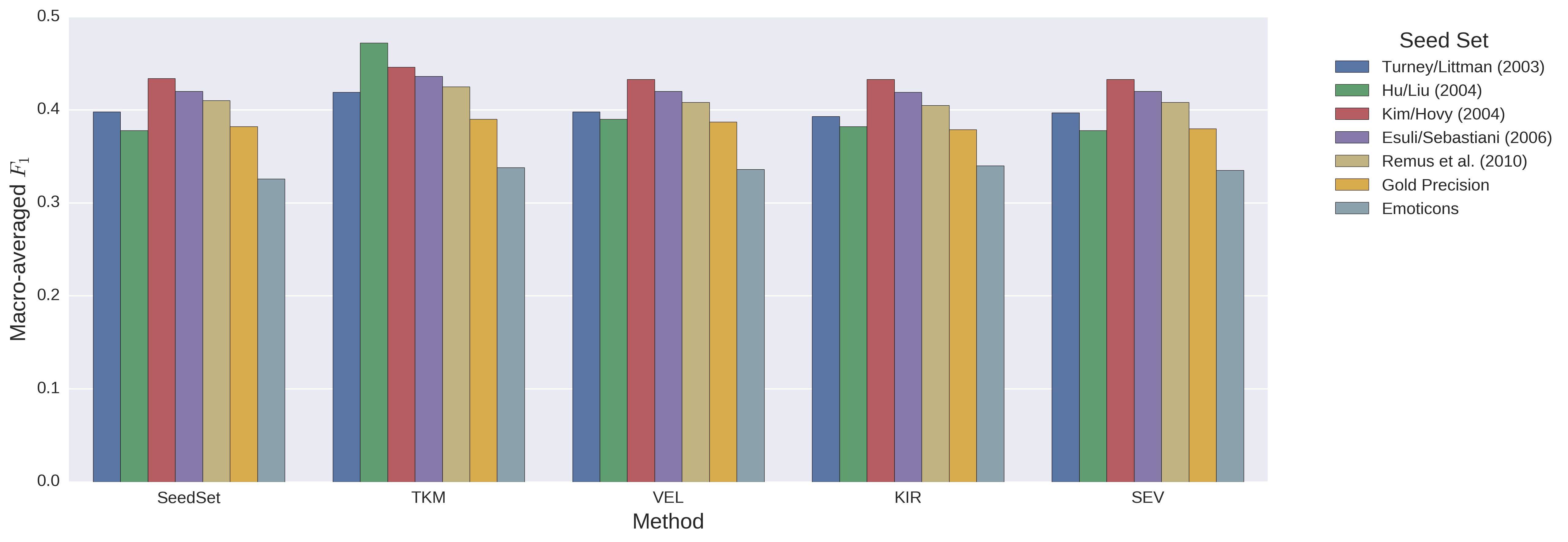}
  \caption{Macro-averaged \F{}-scores of the corpus-based approaches
  with different seed sets.}\label{snt:fig:sent-crp-lex-alt-seeds}
\end{figure}

A slightly different situation is observed for the corpus-based
approaches as shown in Figure~\ref{snt:fig:sent-crp-lex-alt-seeds}.
Except for the method of \newcite{Takamura:05}, all three remaining
methods---\newcite{Velikovich:10}, \newcite{Kiritchenko:14}, and
\newcite{Severyn:15}---show very similar (though not identical)
scores.  Moreover, these scores are also very close to the results
achieved by the respective seed sets without any expansion.  The
primary reasons for this were again the positive bias of the distantly
labeled tweets and the consequently premature stopping of the
expansion.

Following the suggestion of one of the reviewers, we additionally
included two more seed sets in our evaluation: gold precision and
emoticons.  The former list contained just two polar terms---``gut''
(\textit{good}$^+$) and ``schlecht'' (\textit{bad}$^-$)---which showed
an almost perfect precision on the PotTS data
set.\footnote{Unfortunately, we could not include more terms in this
  seed set due to a high lexical ambiguity of other polar words.  Even
  in our proposed prototypical seed list, one of the terms---``gut''
  (\textit{good})---could have another rather rare reading
  (\textit{manor}) when used as a noun.}  The latter seed set
consisted of two regular expressions: one for capturing positive
smileys and another one for matching negative emoticons.  As can be
seen form the figure, these lists, however, could hardly outperform
any of our initially used seed sets.

\section{Analysis of Entries}\label{sec:analysis}

Besides investigating the effects of different hyper-parameters and
seeds, we also decided to have a closer look at the actual results
produced by the tested methods.  For this purpose, we extracted ten
highest scored entries (not counting the seed terms) from each
automatic lexicon and summarized them in
Table~\ref{snt-lex:tbl:top-10}.

\begin{table}[*hbt]
  \begin{center}
    \bgroup \setlength\tabcolsep{0.03\tabcolsep}\scriptsize
    \begin{tabular}{%
        >{\centering\arraybackslash}p{0.02\columnwidth} 
        *{10}{>{\centering\arraybackslash}p{0.095\columnwidth}}} 
      \toprule
      \textbf{Rank} & %
      \textbf{HL} & \textbf{BG} & \textbf{KH} & %
      \textbf{ES} & \textbf{RR}$^{**}_{\textrm{mincut}}$ & %
      \textbf{RR}$_{\textrm{lbl-prop}}$ & %
      \textbf{TKM} & \textbf{VEL} & \textbf{KIR} & %
      \textbf{SEV} \\\midrule
      1 & \ttranslate{perfekt}{perfect} & %
      \ttranslate{flei\ss{}ig}{diligent} &%
      \ttranslate{anr\"uchig}{indecent} &%
      \ttranslate{namenlos}{nameless} &%
      \ttranslate{planieren}{to plane} &%
      \ttranslate{prunkvoll}{splendid} &%
      \ttranslate{Stockfotos}{stock photos} &%
      \ttranslate{Wahl\-kampf\-ge\-schenk}{election gift} &%
      \ttranslate{Suchmaschinen}{search engines} &%
      \ttranslate{Scherwey}{Scherwey}\\

      2 & \ttranslate{musterg\"ultig}{immaculate} & %
      \ttranslate{b\"ose}{evil} &%
      \ttranslate{unecht}{artificial} &%
      \ttranslate{ruhelos}{restless} &%
      \ttranslate{Erdschicht}{stratum} &%
      \ttranslate{sinnlich}{sensual} &%
      \ttranslate{BMKS65}{BMKS65} &%
      \ttranslate{Or\-dens\-ge\-schich\-te}{order history} &%
      \ttranslate{\#gameinsight}{\#gameinsight} &%
      \ttranslate{krebsen}{to crawl}\\

      3 & \ttranslate{vorbildlich}{commendable} & %
      \ttranslate{beispielhaft}{exemplary} &%
      \ttranslate{irregul\"ar}{irregular} &%
      \ttranslate{unbewaffnet}{unarmed} &%
      \ttranslate{gefallen}{please} &%
      \ttranslate{pomp\"os}{ostentatious} &%
      \ttranslate{Ziya}{Ziya} &%
      \ttranslate{Indologica}{Indologica} &%
      \ttranslate{\#androidgames}{\#androidgames} &%
      \ttranslate{kaschieren}{to conceal}\\

      4 & \ttranslate{beispielhaft}{exemplary} & %
      \ttranslate{edel}{noble} &%
      \ttranslate{drittklassig}{third-class} &%
      \ttranslate{interesselos}{indifferent} &%
      \ttranslate{Zeiteinheit}{time unit} &%
      \ttranslate{unappetitlich}{unsavory} &%
      \ttranslate{Shoafoundation}{shoah found.} &%
      \ttranslate{Indologie}{Indology} &%
      \ttranslate{Selamat}{selamat} &%
      \ttranslate{Davis}{Davis}\\

      5 & \ttranslate{exzellent}{excellent} & %
      \ttranslate{t\"uchtig}{proficient} &%
      \ttranslate{sinnlich}{sensual} &%
      \ttranslate{reizlos}{unattractive} &%
      \ttranslate{Derivat}{derivate} &%
      \ttranslate{befehlsgem\"a\ss{}}{as ordered} &%
      \ttranslate{T1199}{T1199} &%
      \ttranslate{Energieverbrauch}{energy consumption} &%
      \ttranslate{Pagi}{Pagi} &%
      \ttranslate{\#Klassiker}{\#classics}\\

      6 & \ttranslate{exzeptionell}{exceptional} & %
      \ttranslate{emsig}{busy} &%
      \ttranslate{unprofessionell}{unprofessional} &%
      \ttranslate{w\"urdelos}{undignified} &%
      \ttranslate{Oberfl\"ache}{surface} &%
      \ttranslate{vierschr\"otig}{beefy} &%
      \ttranslate{Emilay55}{Emilay55} &%
      \ttranslate{Schimmelbildung}{mold formation} &%
      \ttranslate{\#Sparwelt}{\#savingsworld} &%
      \ttranslate{Nationalismus}{nationalism}\\

      7 & \ttranslate{au\ss{}ergew\"ohnlich}{extraordinary} & %
      \ttranslate{eifrig}{eager} &%
      \ttranslate{abgeschlagen}{exhausted} &%
      \ttranslate{absichtslos}{unintentional} &%
      \ttranslate{Essbesteck}{cutlery} &%
      \ttranslate{regelgem\"a\ss}{regularly} &%
      \ttranslate{Eneramo}{Eneramo} &%
      \ttranslate{Hygiene}{hygiene} &%
      \ttranslate{\#Seittest}{\#Seittest} &%
      \ttranslate{Kraftstoff}{fuel}\\

      8 & \ttranslate{au\ss{}erordentlich}{exceptionally} & %
      \ttranslate{arbeitsam}{hardworking} &%
      \ttranslate{gef\"allig}{pleasing} &%
      \ttranslate{ereignislos}{uneventful} &%
      \ttranslate{abl\"osen}{to displace} &%
      \ttranslate{wahrheitsgem\"a\ss}{true} &%
      \ttranslate{GotzeID}{GotzeID} &%
      \ttranslate{wasserd}{waterp} &%
      \ttranslate{Gameinsight}{Gameinsight} &%
      \ttranslate{inaktiv}{idle}\\

      9 & \ttranslate{viertklassig}{fourth-class} & %
      \ttranslate{musterg\"ultig}{exemplary} &%
      \ttranslate{musterg\"ultig}{exemplary} &%
      \ttranslate{regellos}{irregular} &%
      \ttranslate{Musikveranstaltung}{music event} &%
      \ttranslate{fettig}{greasy} &%
      \ttranslate{BSH65}{BSH65} &%
      \ttranslate{heizkostensparen}{saving heating costs} &%
      \ttranslate{\#ipadgames}{\#ipadgames} &%
      \ttranslate{8DD}{8DD}\\

      10 & \ttranslate{sinnreich}{ingenious} & %
      \ttranslate{vorbildlich}{commendable} &%
      \ttranslate{unrecht}{wrong} &%
      \ttranslate{fehlerfrei}{accurate} &%
      \ttranslate{Gebrechen}{afflictions} &%
      \ttranslate{lumpig}{shabby} &%
      \ttranslate{Saymak.}{Saymak.} &%
      \ttranslate{Re\-fe\-renz\-ar\-chi\-tek\-tu\-ren}{reference architectures} &%
      \ttranslate{Fitnesstraining}{fitness training} &%
      \ttranslate{Mailadresse}{mail address}\\\bottomrule
    \end{tabular}
    \egroup
    \caption{Top ten polar terms produced by the automatic methods.\\
      {\small ** -- the min-cut method of \newcite{Rao:09} returns an
        unsorted set}}
    \label{snt-lex:tbl:top-10}
  \end{center}
\end{table}

As can be seen from the table, the approaches of \newcite{Hu:04},
\newcite{Blair-Goldensohn:08}, \newcite{Kim:04}, as well as the
label-propagation algorithm of \newcite{Rao:09} produce almost perfect
polarity lists.  The \textsc{SentiWordNet} approach of
\newcite{Esuli:06c}, however, already features some spurious terms
(e.g., ``absichtslos'' \emph{unintentional}) among its top-scored
entries.  Finally, the min-cut approach of \newcite{Rao:09} returns a
set of mainly objective terms, which, however, is rather due to the
fact that this method performs a cluster-like partitioning of the
lexical graph without ranking the words assigned to a cluster.

An opposite situation is observed for the corpus-based systems: The
top-scoring polarity lists returned by these approaches not only
include many apparently objective terms but are also difficult to
interpret in general, as they contain a substantial number of slang
and advertising terms (e.g., ``BMKS65'', ``\#gameinsight'',
``\#androidgames'' etc.).  This again supports the hypothesis that an
extreme content noisiness of the input domain might pose considerable
difficulties to sentiment lexicon generation methods.

\section{Conclusions and Future Work}\label{sec:concl}

Based on the above observations and our experiments, we can formulate
the main conclusions that we come to in this paper as follows:
\begin{itemize}
\item semi-automatic translations of common English polarity lists
  notably outperform automatic SLG approaches that are applied
  directly to non-English data;
\item despite their allegedly worse ability to accommodate new
  domains, dictionary-based methods are still superior to corpus-based
  systems (at least in terms of the proposed intrinsic evaluation),
  provided that a sufficiently big lexical taxonomy exists for the
  target language;
\item a potential weakness of the dictionary-based algorithms,
  however, is their susceptibility to different hyper-parameter
  settings and the size and composition of the initial seed sets;
\item nevertheless, the effect of the seed sets might be even stronger
  for the corpus-based approaches which rely on distant supervision,
  if the resulting noisy labeled training set becomes highly
  unbalanced.
\end{itemize}
In this respect, there appears to be a great need for a corpus-based
method which can both benefit from in-domain data and be resistant to
non-balanced training sets; and we are, in fact, currently working on
such an algorithm.  By taking advantage of the recent advances in deep
learning and distributional semantics, we aim to show an efficient way
of getting suitable vector representations for polar terms and
generating high-quality sentiment lexicons from these automatically
learned vectors.

\section*{Acknowledgments}\label{sec:ackn}
We thank the anonymous reviewers for their suggestions and comments.

\bibliographystyle{acl}
\bibliography{bibliography}
\end{document}